\documentclass[letterpaper]{article} % DO NOT CHANGE THIS
\usepackage{aaai2027}  % DO NOT CHANGE THIS
\usepackage[hyphens]{url}  % DO NOT CHANGE THIS
\usepackage{graphicx} % DO NOT CHANGE THIS
\usepackage[numbers, sort&compress]{natbib}  % 数字引用，支持[1-3]压缩
\usepackage{caption} % DO NOT CHANGE THIS AND DO NOT ADD ANY OPTIONS TO IT
\usepackage{algorithm}
\usepackage{algorithmic}

\usepackage{newfloat}
\usepackage{listings}
\DeclareCaptionStyle{ruled}{labelfont=normalfont,labelsep=colon,strut=off} % DO NOT CHANGE THIS
\floatstyle{ruled}
\newfloat{listing}{tb}{lst}{}
\floatname{listing}{Listing}

\usepackage{booktabs}

\usepackage{amssymb}
\usepackage{makecell}

\usepackage[colorlinks=true, linkcolor=blue, citecolor=blue, urlcolor=blue]{hyperref}

\usepackage{orcidlink}

\nocopyright

\title{LIBAD: A Multimodal Anomaly Detection Benchmark for Li-Ion Battery Electrode Manufacturing}

\author{
    Wenbo Sui\textsuperscript{\rm 1,2}\corresponding\orcidlink{0000-0001-6437-4456},
    Daniel Lichau\textsuperscript{\rm 1},
    Harold Phelippeau\textsuperscript{\rm 1}\orcidlink{0009-0000-7503-3331},
    Zhao Liu\textsuperscript{\rm 3}\orcidlink{0000-0003-0370-2406}
}

\affiliations{
    \textsuperscript{\rm 1}Thermo Fisher Scientific, 39 Rue d'Armagnac, Bordeaux, 33800, France\\
    \textsuperscript{\rm 2}DTU Energy, Technical University of Denmark, Fysikvej 310, Kgs.\ Lyngby, 2800, Denmark\\
    \textsuperscript{\rm 3}Thermo Fisher Scientific, 5350 NE Dawson Creek Dr, Hillsboro, OR 97124, USA\\
    wenbo.sui@thermofisher.com
}

\begin{document}

\maketitle

\begin{abstract}
Multimodal industrial anomaly detection has largely focused on discrete products using strongly correlated RGB and 3D observations, leaving continuous process manufacturing and weakly correlated sensing modalities underexplored. We introduce LIBAD, the first multimodal anomaly detection benchmark for Li-ion battery electrode manufacturing. Collected from real roll-to-roll production lines, LIBAD provides aligned double-sided visible-light imaging, high-resolution X-ray radiography, and inline-compatible low-resolution X-ray radiography. Electrode patches in LIBAD exhibit highly homogeneous material appearance, while defect evidence can be strong in one modality but weak or absent in another, resulting in pronounced cross-modal anomaly inconsistency. Benchmarks of representative methods under the inline-compatible visible-light and low-resolution X-ray setting exhibit limited transferability and consistently high false-positive rates. We therefore propose DA-Core, a memory-based method that jointly considers feature-space coverage and local density of normal features during coreset selection, allowing compact memory banks to better preserve fine-grained normal variations. With a coreset ratio of 0.05, DA-Core reduces FPR95 from 60.4\% to 54.3\% compared with standard farthest point sampling. At this ratio, DA-Core also outperforms the best standard coreset result (obtained at 0.20) while reducing inference time by 43.9\%. These results suggest that both the data distribution of normal features and the modality relationship itself require explicit consideration when designing anomaly detection methods for process manufacturing.
\begin{links}
    \link{Code\&Datasets}{https://github.com/evenrose/LIBAD}
    % \link{Datasets}{https://github.com/evenrose/LIBAD}
    % \link{Extended version}{https://aaai.org/example/extended-version}
\end{links}
\end{abstract}

\section{Introduction}

With the rapid development of smart manufacturing, industrial anomaly detection (IAD) has become an important tool for quality control, yield improvement, and process monitoring. Recent multimodal methods improve inspection reliability by combining complementary sensing signals. Most existing studies focus on RGB and 3D observations, where appearance and geometry provide strongly correlated cues for identifying surface defects and structural deformations \cite{liu2024deep,LIN2025103139,li2025survey}.

Despite this progress, existing multimodal IAD benchmarks remain limited in two respects. First, most datasets are built around discrete, object-centric products \cite{bergmann2021mvtec3d,bonfiglioli2022eyecandies,zhu2025real,li2025multi}, leaving continuous process manufacturing, such as roll-to-roll production, largely unexplored. Second, existing benchmarks commonly assume that an anomaly produces corresponding evidence across modalities, which is reasonable for RGB and 3D observations of the same object surface but does not generally hold for sensing modalities that measure different physical properties. In real systems, a defect may be visible in one modality yet weak or absent in another, resulting in inconsistent cross-modal anomaly evidence.

\begin{figure}[t]
\centering
\includegraphics[width=\linewidth]{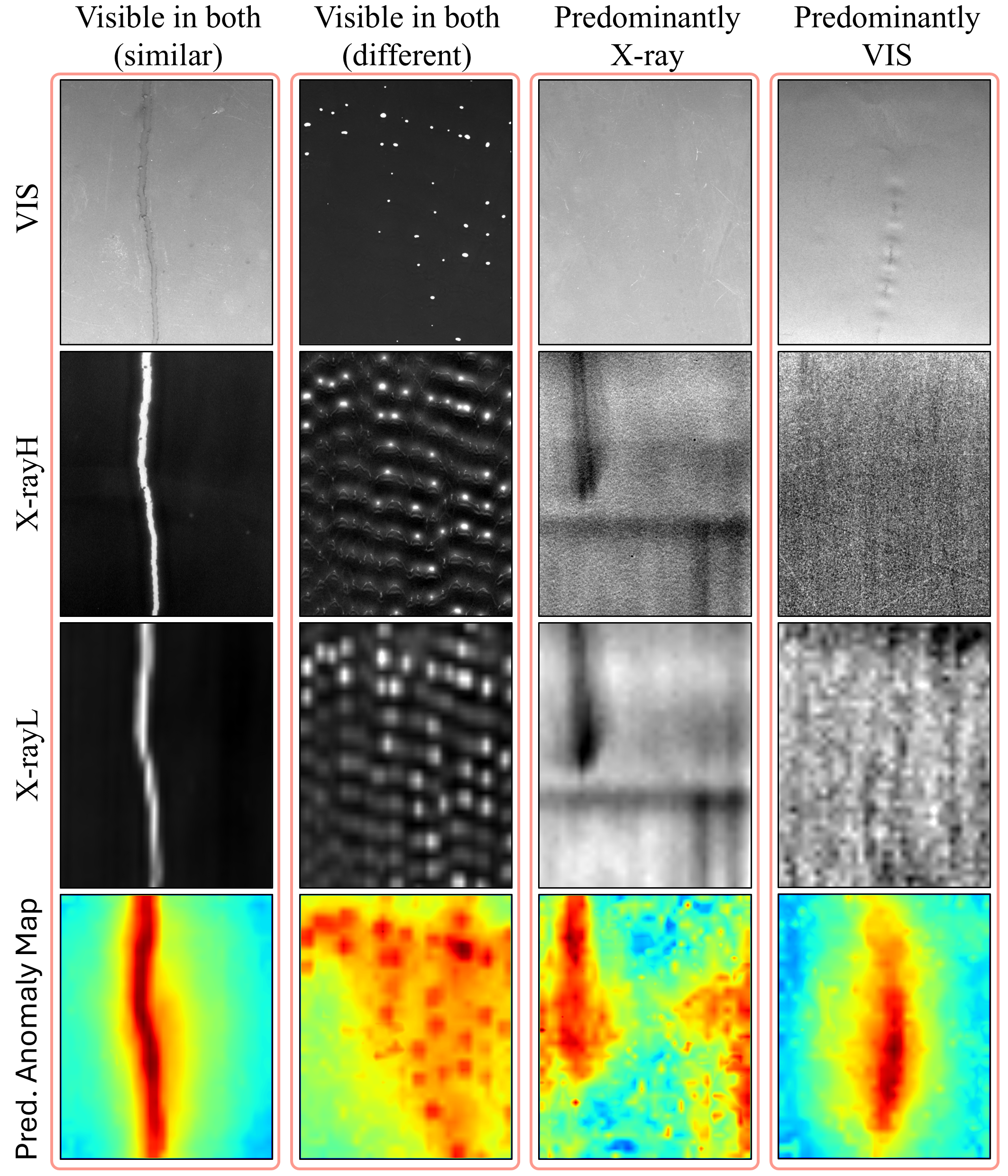}
\caption{Example defects in LIBAD across four visibility patterns. Each row shows VIS, X-rayH, X-rayL, and the predicted anomaly map of DA-Core.}
\label{fig:defect_types}
\end{figure}

To address these limitations, we introduce LIBAD, the first multimodal anomaly detection benchmark for Li-ion battery electrode manufacturing. LIBAD contains 744 electrode samples collected from real roll-to-roll production lines, with three spatially aligned modalities: visible-light grayscale imaging (VIS), high-resolution X-ray radiography (X-rayH), and inline-compatible low-resolution X-ray radiography (X-rayL). These modalities capture complementary surface and internal properties of double-sided coated electrodes, for which subsurface defects cannot be reliably identified through surface inspection alone. 
LIBAD covers 11 real-world defect categories and provides image-level anomaly labels for each electrode patch. Because visible light and X-ray probe surface appearance and bulk density respectively, defects that are prominent in one modality may be weak or entirely absent in the other, as illustrated in Fig.~\ref{fig:defect_types}. LIBAD therefore evaluates whether a method can integrate complementary but potentially conflicting modality evidence, rather than relying on redundant anomaly cues shared across modalities. At the same time, it is designed to address a pressing industrial need: undetected electrode defects can compromise battery safety and performance in electric vehicles and energy storage systems \cite{7795214,Attia2025}.

Our benchmark shows that existing multimodal IAD methods transfer poorly to LIBAD. Popular methods based on cross-modal reconstruction often assume stable semantic correspondence between modalities, causing normal modality differences to be confused with anomalies \cite{costanzino2024multimodal,cfr2026, zhao2026complementary}. Methods designed specifically for RGB--3D inputs additionally rely on appearance--geometry priors that do not generalize to visible-light and X-ray observations \cite{tao2025g2sf, li2025hgcf}. We therefore establish a memory-based baseline that avoids explicitly enforcing cross-modal reconstruction or correspondence. However, conventional farthest point sampling (FPS) \cite{roth2022towards} prioritizes global coverage but ignores local sample density, which is problematic for electrode data with highly concentrated normal patterns and subtle intra-class variations. By favoring sparse regions over dense normal clusters, FPS tends to produce a memory bank that under-represents the most common normal patterns, leading to unnecessary false positives during nearest-neighbor scoring. We therefore propose DA-Core, a density-aware coreset selection strategy that combines global coverage with local feature density during memory construction, reducing false-positive detections while maintaining a compact memory bank.

Our contributions are summarized as follows:
\begin{itemize}
    \item We introduce LIBAD, the first multimodal IAD benchmark for Li-ion battery electrode manufacturing, comprising 744 real-world samples, three spatially aligned sensing modalities, and 11 defect categories collected from a roll-to-roll production line.

    \item We characterize cross-modal anomaly inconsistency in electrode inspection, where defect evidence is often modality-selective rather than shared.

    \item We benchmark representative multimodal IAD methods under a unified evaluation protocol and analyze the limitations of existing fusion, reconstruction, and modality-specific designs.

    \item We propose DA-Core, a density-aware coreset selection method that constructs a more representative memory bank from non-uniform distributions of normal features and reduces false-positive detections.
\end{itemize}

\section{Related Work}

\subsection{Industrial Anomaly Detection Benchmarks}
Public benchmarks have played a central role in the development of IAD methods. MVTec AD \cite{bergmann2019mvtec} established a widely used benchmark with multiple object and texture categories and pixel-level annotations. MVTec 3D-AD \cite{bergmann2021mvtec3d} further introduced paired RGB and 3D point cloud data, laying an important foundation for multimodal industrial anomaly detection based on appearance and geometry. Recent benchmarks extend toward richer sensing: Real-IAD D$^3$ \cite{zhu2025real} provides aligned 2D, pseudo-3D, and 3D data for real-world industrial anomaly detection, while MulSen-AD \cite{li2025multi} unifies RGB cameras, laser scanners, and infrared thermography to capture appearance, geometry, and internal properties. Despite these advances, existing benchmarks mainly focus on inspection of discrete, object-centric products, leaving process manufacturing underexplored. LIBAD complements them with spatially aligned visible-light and X-ray data from a real roll-to-roll electrode production line.

\subsection{Multimodal Industrial Anomaly Detection}
Most existing multimodal IAD methods are developed for RGB and 3D data, where appearance and geometry provide strongly related descriptions of the same object. M3DM \cite{wang2023multimodal} introduces patch-wise contrastive learning to align RGB and 3D features from the same spatial location, while separating features from different locations within the same modality. CFM \cite{costanzino2024multimodal}, CFR \cite{cfr2026}, and CPMAD \cite{zhao2026complementary} instead learn cross-modal feature mappings and detect anomalies through reconstruction residuals. However, in LIBAD's visible-light and X-ray setup, the weak correspondence between modalities can cause reconstruction failure even for normal samples. G$^{2}$SF \cite{tao2025g2sf} exploits geometry-guided structural fusion, and HGCF \cite{li2025hgcf} further models hierarchical color-geometry interactions. Both methods rely on RGB-3D priors that are not available in visible-light and X-ray inspection. These limitations suggest that LIBAD requires methods that can better handle weak and inconsistent multimodal evidence.

\subsection{Memory-based Anomaly Detection and Coreset Selection}
SPADE \cite{cohen2020sub} and PaDiM \cite{defard2021padim} model normality with stored normal features and estimate anomalies by nearest-neighbor matching. PatchCore \cite{roth2022towards} further introduces coreset selection to reduce memory size while preserving feature-space coverage. Recent extensions include zero-shot/few-shot retrieval methods like DMMGNet \cite{luo2024dmmgnet}, ReMP-AD \cite{ma2025remp}, as well as multimodal adaptations such as MulSen-TripleAD \cite{li2025multi} with a decision gating unit, D$^3$-Memory \cite{zhu2025real} with channel-spatial swapping. 
However, existing coreset selection strategies mainly optimize global coverage, which tends to under-represent dense local structures in the normal feature distribution.

\section{The LIBAD Dataset}

\subsection{Dataset Construction}

Electrode manufacturing produces continuous webs at high throughput, where defects are often sparse, making full-web annotation too expensive. Thus, we formulate LIBAD as an image-level anomaly detection dataset with physically cropped electrode patches as the basic unit. This design offers three practical advantages: (1) it makes data collection and registration more tractable; (2) it allows the dataset to easily scale to new defect types or process conditions; (3) it enables highly parallel inference for inline evaluation.

\subsubsection{Sample Collection and Annotation}

All electrode patches are collected from real roll-to-roll Li-ion battery electrode production lines (see Appendix), covering multiple cathode and anode compositions and production conditions. Candidate defective regions are identified through offline inspection based on visible appearance and measurable physical deviations. Once a candidate defective region is located on the electrode web, a physical sample slightly larger than $50 \times 40$ mm is cut from the corresponding position.
Initial defect labels are assigned by experienced process engineers from the lines and reviewed with the assistance of battery researchers. After multimodal image acquisition, all labels are re-examined using both visible-light and X-ray observations by joint experts. Each sampled region is assigned a single defect label according to its dominant defect pattern.

A major concern in sample collection is that anomalous and normal samples may differ not only in defect patterns, but also in material composition, production batch, or process settings. To reduce these confounding effects, normal patches are collected near the defective samples or under the same manufacturing conditions whenever possible, so they are naturally grouped by the associated defect collection conditions. These groups are used for dataset statistics and stratified splitting, but all selected normal samples are pooled together during training to construct a single model. This design encourages the benchmark to focus on defect-related evidence rather than unrelated material, batch, or process differences.

\subsubsection{Multimodal Data Acquisition}

Each electrode patch is characterized using visible-light imaging and two X-ray radiography settings. Visible-light images (VIS) are acquired under coaxial illumination, producing double-sided grayscale surface images of the coated electrode with an effective pixel size of approximately $0.048$ mm/pixel. 
High-resolution X-ray radiographs (X-rayH) are acquired using a micro-CT system (45 kV, 85 $\mu$A), with an effective pixel size of approximately $0.045$ mm/pixel and providing internal structure and density information at a resolution comparable to VIS. This setting is used for analyzing defects with weak or missing surface signatures, but its relatively long exposure time makes it only suitable for offline inspection. Although X-rayH is available for every sample, we use it only for label annotation to align with the practical inline VIS+X-rayL scenario. Low-resolution X-ray radiographs (X-rayL) are acquired using an inline X-ray system for roll-to-roll production. The resulting images have an effective pixel size of approximately $1.1$ mm/pixel, trading spatial resolution for acquisition speed while retaining sensitivity to density variations (see Appendix for detailed imaging-system configurations). In particular, visible-light imaging and inline X-ray sensing can be integrated into the same inspection line for deployment, with lightweight spatial calibration for aligned patch-wise inference.

\subsubsection{Multimodal Registration}

We first crop and orient each sample from the raw images to isolate the electrode region. All images are then registered to one side of the visible-light grayscale image, which serves as the fixed reference. Registration is performed by optimizing an affine transform to maximize a correlation-ratio similarity measure between the fixed and moving images \cite{Roche1998}. For a fixed image $X$ and a moving image $Y$ over the overlapping region, with paired intensity samples $\{(x_i, y_i)\}_{i=1}^{N}$, the correlation ratio is defined as:
\begin{equation}
\eta(Y|X)
=
\frac{
\sum_{k} n_k \, (\bar{y}_k - \bar{y})^2
}{
\sum_{i=1}^{N} (y_i - \bar{y})^2
},
\end{equation}
where pixels are grouped by their intensity $x_i = k$ in the fixed image, $\bar{y}_k$ is the mean intensity of the corresponding pixels in the moving image, $\bar{y}$ is the global mean, and $n_k$ is the number of pixels in group $k$. The numerator measures the between-group variance of $Y$ explained by grouping according to $X$, while the denominator is the total variance of $Y$. A higher ratio indicates stronger cross-modal correspondence, meaning the two images are better aligned. This measure captures functional intensity dependence without assuming linearity, making it well suited for this kind of registration.

We use a coarse-to-fine derivative-free line search optimizer to find the affine parameters (without shear) that maximize $\eta(Y|X)$. Registration quality is verified using the final correlation-ratio score, and we set the acceptance threshold to 0.65 based on empirical separation between aligned and misaligned cases. All samples pass this threshold and quantitative evaluation on 20 randomly selected pairs yields a mean reprojection error of 2.1 \textpm\ 1.5 pixels (0.08 mm), confirming reliable alignment. After registration, the aligned $50 \times 40$ mm physical region is resampled to a common $1280 \times 1024$ grid using bicubic interpolation which provides a sufficiently fine coordinate grid for alignment. Before feature extraction, all modality inputs are resized to $640 \times 512$ pixels.

\subsection{Dataset Statistics and Characteristics}

LIBAD contains 744 electrode patches, including 361 normal and 383 anomalous patches from 11 real-world defect categories as shown in Table~\ref{tab:dataset_statistics}. Following MVTec 3D-AD \cite{bergmann2021mvtec3d}, training and validation sets contain normal samples only, while the test set contains both normal and anomalous samples. To mitigate evaluation bias caused by random partitioning, we provide 10 official splits generated by randomly partitioning normal samples within each defect group. In each split, approximately 55\% of the normal patches are used for training, 15\% for validation, and the remaining 30\% for testing. The resulting numbers of samples in each subset are summarized in Table~\ref{tab:dataset_statistics}. Normal samples are grouped by their matched defect collections for stratified splitting and are pooled across all groups to train a single model. Compared with MVTec 3D-AD, LIBAD retains a larger proportion of normal samples in the test set to support the evaluation of false-positive behavior. 

\begin{table}[t]
\centering
\small
\setlength{\tabcolsep}{7pt}
\begin{tabular}{lccccc}
\toprule
Group
& \# Train
& \# Val
& \makecell{\# Test\\(normal)}
& \makecell{\# Test\\(anomalous)} \\
\midrule
Wrinkling  & 21 & 6 & 11 & 38 \\
Particle   & 17 & 5 & 9  & 34 \\
Pit        & 19 & 5 & 11 & 37 \\
Unevenness & 19 & 5 & 10 & 31 \\
Barefoil   & 16 & 5 & 9  & 30 \\
Scratch    & 20 & 6 & 11 & 35 \\
Polarity   & 19 & 5 & 10 & 35 \\
Debonding  & 20 & 6 & 12 & 20 \\
Crack      & 19 & 5 & 11 & 35 \\
Streak     & 18 & 5 & 10 & 35 \\
Pinhole    & 9  & 2 & 5  & 53 \\
\midrule
Total      & 197 & 55 & 109 & 383 \\
\bottomrule
\end{tabular}
\caption{Dataset split statistics of LIBAD.}
\label{tab:dataset_statistics}
\end{table}

Beyond the class distribution, LIBAD exhibits modality-dependent defect visibility. As shown in Fig.~\ref{fig:defect_types}, some defects are visible in both VIS and X-ray with similar appearance, while others are visible in both imaging settings but show different patterns. Some defects are predominantly detectable in X-ray, indicating internal or density-related abnormalities with weak surface evidence, whereas others are mainly visible in VIS and become weak or ambiguous in X-ray. This property makes LIBAD different from MVTec 3D-AD or Real-IAD D$^3$ where anomalous evidence is largely shared across modalities. Accurately defining which defects are visible in which modality is difficult, given the limited physical prior knowledge of defects and their complex appearance. We provide an analysis using mutual information scores between registered modalities, supplemented by manual inspection (see Appendix).
Another characteristic of LIBAD is the homogeneous appearance of electrode patches. Unlike other datasets where the target object occupies the center part of the image and often contains rich semantic structures, each LIBAD patch is entirely covered by electrode material, making normal regions from different locations visually similar.

\section{Benchmark Protocol}

\subsection{Task Definition}

LIBAD follows the standard unsupervised anomaly detection setting: normal samples are used for training and validation, reflecting the practical scenario where defects are unseen during training but normality is well-defined. During evaluation, each test sample is assigned an image-level anomaly score and classified as normal or anomalous. We adopt this sample-level evaluation because defect boundaries can be ambiguous across imaging modalities, while patch rejection is sufficient for the considered inspection setting.
We use VIS and X-rayL as the main benchmark modalities because they represent the most practical inline inspection setting, combining surface appearance with density-sensitive X-ray measurement. We report unimodal results on VIS and X-rayL as references, and use VIS+X-rayL as the main multimodal setting. X-rayH is not used in the benchmark. It is reserved for label annotation and defect analysis due to its offline acquisition setting.

\subsection{Evaluation Metrics}

We report image-level AUROC, AUPR, F1-max, and FPR95 for anomaly detection. AUROC and AUPR are widely used threshold-independent metrics that measure the ranking quality of anomaly scores. However, industrial inspection systems must eventually operate with a decision threshold, and strong ranking performance does not necessarily imply a low false alarm rate in deployment. We therefore emphasize FPR95, which measures the false positive rate when the true positive rate reaches 95\%. This metric is particularly important for electrode inspection, where excessive false alarms can make an automatic inspection system impractical.
We also report inference speed as an efficiency metric, since inline inspection requires both reliable detection and efficient scoring. All results are reported as the mean and standard deviation over the 10 official random splits.

\subsection{Baselines}

We benchmark recent representative methods in IAD, including PatchCore \cite{roth2022towards}, M3DM \cite{wang2023multimodal}, CFM \cite{costanzino2024multimodal}, G$^{2}$SF \cite{tao2025g2sf}, and CFR \cite{cfr2026}.
For methods originally designed for RGB--3D inputs, we adapt their modality branches to VIS and X-rayL, since LIBAD does not provide point-cloud observations. To enable a fair comparison, we replace their original DINOv2/DINOv1 backbones with DINOv3 \cite{simeoni2025dinov3}, which yields consistent improvements (see Appendix).

\section{Method}

\begin{figure}[ht]
\centering
\includegraphics[width=\columnwidth]{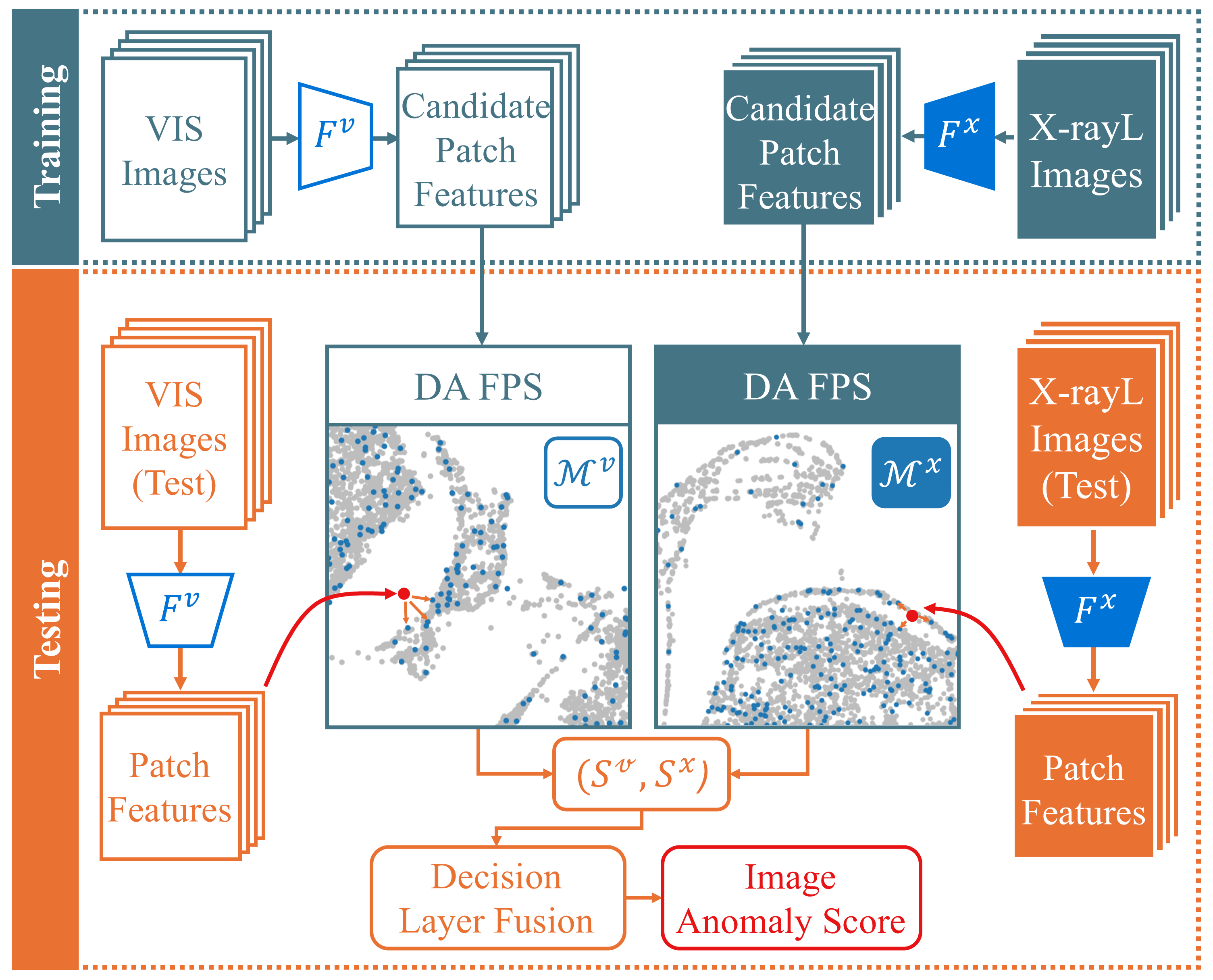}
\caption{Overview of DA-Core.}
\label{fig:model}
\end{figure}

We propose DA-Core, a memory-based anomaly detection method for LIBAD. DA-Core improves the standard PatchCore-style pipeline by introducing density-aware coreset selection during memory bank construction. This design suppresses false positives caused by under-represented dense normal patterns. The method consists of four steps as shown in Fig.~\ref{fig:model}. First, frozen encoders extract patch-level features from visible-light and X-ray images. Second, normal patch features are stored in modality-specific memory banks and selected by density-aware farthest point sampling (DA FPS). Third, test samples are scored by nearest-neighbor matching with reweighted image-level anomaly scores. Finally, a One-Class Support Vector Machine (OCSVM) late-fusion module combines VIS and X-rayL scores into the final anomaly score, following the strategy used in M3DM.

\subsection{Feature Extraction}

DA-Core uses frozen DINOv3 ViT-S/16 \cite{simeoni2025dinov3} to extract patch-level features. For visible-light imaging, the two electrode sides (a and b) are processed by the same encoder but treated as independent inputs. Their patch tokens are inserted into a shared VIS memory candidate set:
\begin{equation}\label{zv}
    \mathcal{Z}^{v}
    =
    \bigcup_i
    \left(
    F_{v}(x_i^{v,a}) \cup F_{v}(x_i^{v,b})
    \right).
\end{equation}
For X-rayL, patch tokens are extracted by a separate encoder and form the X-rayL memory candidate set:
\begin{equation}\label{zx}
    \mathcal{Z}^{x}
    =
    \bigcup_i
    F_{x}(x_i^{x}).
\end{equation}

\subsection{Density-aware Coreset Construction}

For each modality $m \in \{v,x\}$, let
$\mathcal{Z}^{m}$ denote the full candidate set. The goal of coreset selection is to construct a compact memory bank $\mathcal{M}^{m}\subset\mathcal{Z}^{m}$ while preserving representative normal patterns. Standard farthest point sampling selects the next feature according to its distance to the current memory bank \cite{roth2022towards}:
\begin{equation}
    r_i^{\mathcal{M}}
    =
    \min_{u \in \mathcal{M}^{m}} d(z_i,u),
\end{equation}
\begin{equation}
    z^{*}
    =
    \arg\max_{z_i \in \mathcal{Z}^{m}\setminus \mathcal{M}^{m}}
    r_i^{\mathcal{M}},
\end{equation}
where $d(\cdot,\cdot)$ denotes the feature-space distance. This strategy provides broad coverage of the feature space of normal patches, but considers only geometric distance and ignores the empirical density of normal features. Dense regions often correspond to common electrode coating textures and contain subtle but valid normal variations. Representing such regions with too few memory features can increase nearest-neighbor distances for hard normal samples.

To account for the local distribution of normal features, DA-Core estimates the density of each feature before coreset selection. For a feature $z_i$, we compute its $k$ nearest neighbors in the full normal feature set and define
\begin{equation}
    \rho_i
    =
    \sum_{z_j \in \mathcal{N}_k(z_i)}
    \exp
    \left(
    -
    \left(
    \frac{d(z_i,z_j)}{\tau}
    \right)^2
    \right),
\end{equation}
where $\mathcal{N}_k(z_i)$ denotes the $k$ nearest neighbors of $z_i$, and $\tau$ is the median $k$-nearest-neighbor distance of sampled normal features. The density score is log-transformed as
\begin{equation}
    \hat{\rho}_i
    =
    \log(1+\rho_i),
\end{equation}
followed by quantile normalization to obtain $\tilde{\rho}_i$.

DA-Core combines the coverage distance and local density into a density-aware FPS score:
\begin{equation}
    q_i
    =
    \tilde{r}_i^{\mathcal{M}}
    \left(
    1+\lambda \tilde{\rho}_i
    \right),
\end{equation}
where $\tilde{r}_i^{\mathcal{M}}$ is obtained by Min--Max normalization of the distances to the current memory bank across all remaining candidates, and $\lambda$ controls the contribution of local density. The next memory feature is selected as
\begin{equation}
    z^{*}
    =
    \arg\max_{z_i \in \mathcal{Z}^{m}\setminus \mathcal{M}^{m}}
    q_i.
\end{equation}
The distance term preserves the global coverage behavior of farthest point sampling, while the density term favors representative variations in densely populated normal regions.

\subsection{Anomaly Scoring}

DA-Core follows the nearest-neighbor anomaly scoring strategy and adopts late decision fusion with OCSVM to keep modality-specific evidence before the final decision, similar to M3DM \cite{wang2023multimodal} and CMDIAD \cite{sui2025incomplete}. At inference time, VIS and X-ray feature maps are compared with their corresponding memory banks, producing two modality-level anomaly scores. An OCSVM decision function $C$ is then applied to obtain the final image-level anomaly score:
\begin{equation}
    A(x)
    =
    C
    \left(
    \alpha \psi(F^{v}, \mathcal{M}^{v}),
    \beta \psi(F^{x}, \mathcal{M}^{x})
    \right),
\end{equation}
where $F^{v}$ and $F^{x}$ denote the VIS and X-ray feature maps, $\mathcal{M}^{v}$ and $\mathcal{M}^{x}$ are their corresponding memory banks, and $\alpha$ and $\beta$ are modality scaling factors. The OCSVM is trained using only normal training evidence.

The modality-level score $\psi(F,\mathcal{M})$ is computed from the most anomalous patch feature in a feature map. Specifically,
\begin{equation}
    \psi(F, \mathcal{M}) = d(z^{*}, u^{*}),
\end{equation}
where the most anomalous patch feature $z^{*}$ and its nearest memory feature $u^{*}$ are defined as
% \begin{equation}
%     z^{*}, u^{*}
%     =
%     \arg\max_{z \in F}
%     \min_{u \in \mathcal{M}}
%     d(z,u).
% \end{equation}
\begin{equation}
    z^{*}
    =
    \arg\max_{z \in F}
    \min_{u \in \mathcal{M}}
    d(z,u),
\end{equation}
\begin{equation}
    u^{*}
    =
    \arg\min_{u \in \mathcal{M}}
    d(z^{*},u).
\end{equation}
This formulation follows the standard PatchCore-style scoring, where the image-level anomaly score is the maximum nearest-neighbor distance across all patches. For VIS, side a and b are scored independently using the same VIS memory bank, and the larger score is used as $\psi(F^{v}, \mathcal{M}^{v})$. For X-ray, the score is computed directly from the X-ray feature map and memory bank.

\section{Experiments}

\begin{table*}[t]
\centering
\small
\begin{tabular}{llccccc}
\toprule
Method & Input & FPR95 $\downarrow$ & AUROC $\uparrow$ & AUPR $\uparrow$ & F1-max $\uparrow$ & Time $\downarrow$ \\
\midrule
\multicolumn{7}{c}{Unimodal settings} \\
PatchCore \cite{roth2022towards} & VIS & 0.676 \textpm\ 0.036 & 0.842 \textpm\ 0.012 & 0.951 \textpm\ 0.005 & 0.891 \textpm\ 0.004 & 8.63 ms \\
PatchCore \cite{roth2022towards} & X-rayL & 0.899 \textpm\ 0.025 & 0.712 \textpm\ 0.016 & 0.912 \textpm\ 0.006 & 0.877 \textpm\ 0.002 & 4.10 ms \\
DA-Core & VIS & 0.645 \textpm\ 0.041 & 0.845 \textpm\ 0.016 & 0.951 \textpm\ 0.006 & 0.896 \textpm\ 0.004 & 8.65 ms \\
DA-Core & X-rayL & 0.918 \textpm\ 0.035 & 0.706 \textpm\ 0.013 & 0.909 \textpm\ 0.006 & 0.876 \textpm\ 0.001 & 4.08 ms \\
\midrule
\multicolumn{7}{c}{Multimodal settings} \\
PatchCore \cite{roth2022towards} & VIS+X-rayL & 0.604 \textpm\ 0.039 & 0.865 \textpm\ 0.013 & 0.957 \textpm\ 0.005 & 0.900 \textpm\ 0.005 & 12.99 ms \\
M3DM \cite{wang2023multimodal} & VIS+X-rayL & 0.616 \textpm\ 0.032 & 0.862 \textpm\ 0.011 & 0.957 \textpm\ 0.004 & 0.898 \textpm\ 0.003 & 24.54 ms \\
CFM \cite{costanzino2024multimodal} & VIS+X-rayL & 0.618 \textpm\ 0.037 & 0.834 \textpm\ 0.017 & 0.947 \textpm\ 0.006 & 0.897 \textpm\ 0.004 & 7.24 ms \\
G$^{2}$SF \cite{tao2025g2sf} & VIS+X-rayL & 0.595 \textpm\ 0.048 & 0.852 \textpm\ 0.018 & 0.953 \textpm\ 0.007 & 0.903 \textpm\ 0.005 & 19.74 ms \\
CFR \cite{cfr2026} & VIS+X-rayL & 0.681 \textpm\ 0.037 & 0.833 \textpm\ 0.015 & 0.948 \textpm\ 0.006 & 0.892 \textpm\ 0.005 & 10.61 ms \\
\midrule
DA-Core & VIS+X-rayL & \textbf{0.543 \textpm\ 0.057} & \textbf{0.867 \textpm\ 0.016} & \textbf{0.957 \textpm\ 0.006} & \textbf{0.906 \textpm\ 0.006} & 13.12 ms \\
\bottomrule
\end{tabular}
\caption{Image-level anomaly detection results on LIBAD. Results are reported as mean and standard deviation over 10 official splits. Inference time is measured per sample on an NVIDIA RTX PRO 6000 Blackwell GPU (300W) under the same evaluation setting, excluding data loading (see Appendix for full hardware and software specifications).}
\label{tab:main_results}
\end{table*}

\subsection{Anomaly Detection on LIBAD}

Table~\ref{tab:main_results} reports image-level anomaly detection results on LIBAD using VIS and X-rayL across the 10 official splits.
In the unimodal setting, VIS substantially outperforms X-rayL, with PatchCore achieving 84.2\% vs.\ 71.2\% AUROC and 67.6\% vs.\ 89.9\% FPR95. This indicates that low-resolution X-ray alone is less reliable for inspection. DA-Core reduces VIS FPR95 from 67.6\% to 64.5\%, but shows no improvement on X-rayL alone, consistent with the expectation that density-aware FPS is beneficial when the feature distribution of normal patches contains rich within-class variations.

Combining VIS and X-rayL improves over either modality alone: PatchCore with VIS+X-rayL reduces FPR95 to 60.4\% and improves AUROC to 86.5\%. However, the adapted multimodal baselines show limited transferability to LIBAD, with FPR95 ranging from 59.5\% (G$^{2}$SF) to 69.7\% (CFR), barely improving over PatchCore under the VIS+X-rayL setting (60.4\%). The results are consistent with two likely explanations. For reconstruction-based methods such as CFM and CFR, the weak correlation between visible-light and X-ray images causes large reconstruction residuals even for normal samples, making anomaly discrimination unreliable. For methods relying on RGB-3D priors such as M3DM and G$^{2}$SF, the appearance-geometry consistency assumption is violated in the present multimodal setting. These results suggest that LIBAD poses challenges beyond those addressed by existing multimodal IAD methods.

DA-Core achieves the best overall performance, with 54.3\% FPR95, 86.7\% AUROC, 95.7\% AUPR, and 90.6\% F1-max. Compared with multimodal PatchCore, DA-Core reduces FPR95 by 6.1 percentage points at comparable inference speed (13.12 ms vs.\ 12.99 ms per electrode sample). It also outperforms the best baseline (G$^{2}$SF) by 5.2 percentage points in FPR95 and runs 1.5$\times$ faster with similar memory bank structure (13.12 ms vs.\ 19.74 ms). The improvement is largely driven by false-positive suppression, which reflects the design goal of DA-Core.

\subsection{Ablation Study}

\begin{figure*}[ht]
\centering
\includegraphics[width=0.90\textwidth]{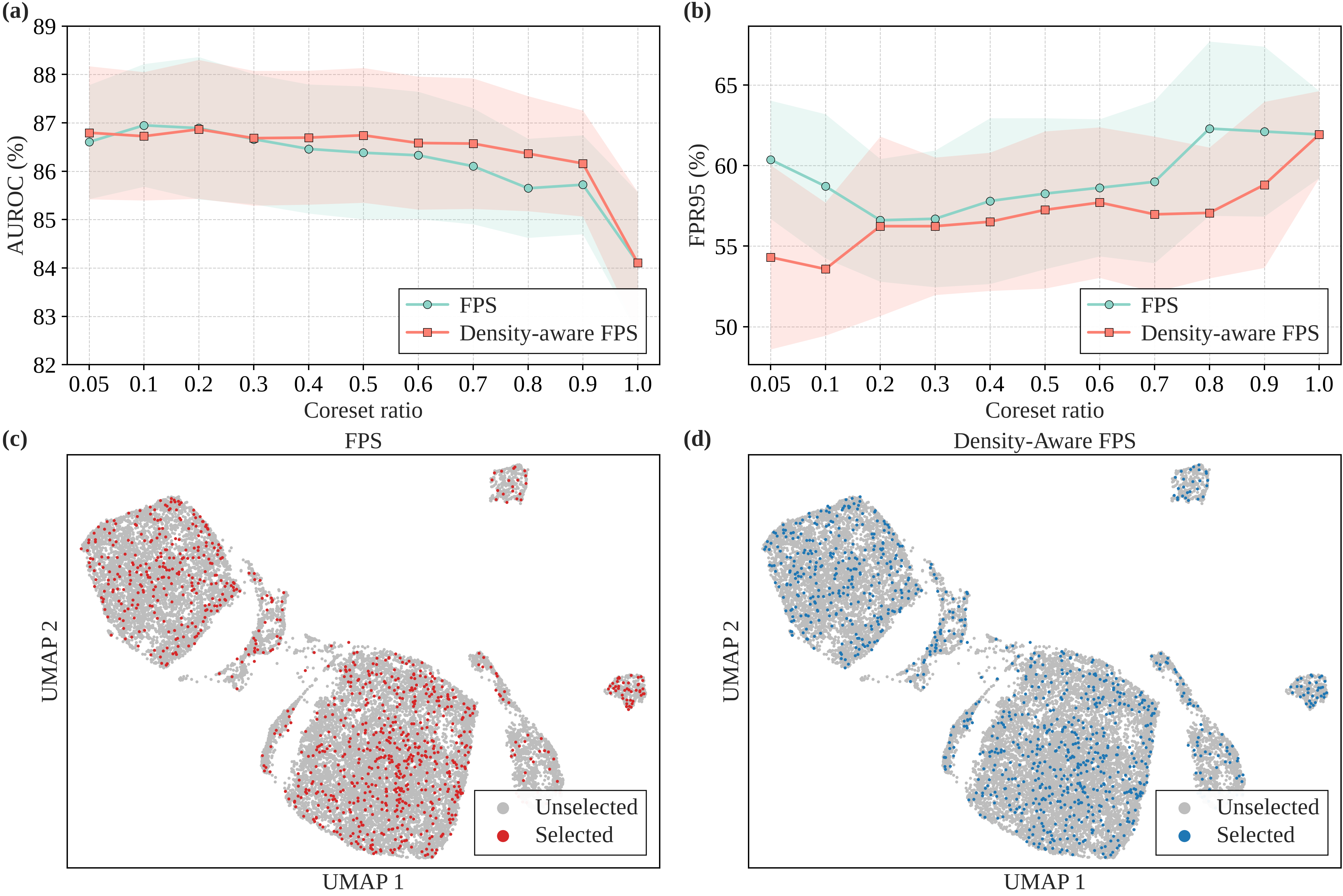}
\caption{Quantitative and qualitative comparison of standard FPS and density-aware FPS. (a) FPR95 and (b) AUROC under different coreset ratios, reported as mean and standard deviation. (c, d) UMAP visualizations of randomly sampled 20,000 VIS normal patches and the coresets selected by standard FPS and density-aware FPS, respectively.}
\label{fig:coreset_ratio_and_selection_combined}
\end{figure*}

\subsubsection{Effect of density-aware coreset selection.}
We first study the effect of density-aware coreset selection under different memory budgets, keeping all other components fixed. Fig.~\ref{fig:coreset_ratio_and_selection_combined}(a,b) compares standard FPS and density-aware FPS across coreset ratios ranging from 0.05 to 1.0. DA-Core improves the performance significantly under compact memory budgets. With a coreset ratio of 0.05, standard FPS obtains 60.4\% FPR95, while density-aware FPS reduces it to 54.3\%, corresponding to a 6.1 percentage point reduction. A similar trend is observed at a coreset ratio of 0.10, where FPR95 decreases from 58.7\% to 53.6\%. 

Density-aware FPS also reaches strong performance with a smaller memory bank. Using only 5\% of the memory candidates, it achieves 54.3\% FPR95, lower than the best standard FPS result among the tested ratios, 56.6\% at a coreset ratio of 0.20. The UMAP visualizations in Fig.~\ref{fig:coreset_ratio_and_selection_combined}(c,d) further illustrate the difference between the two selection strategies. Standard FPS mainly emphasizes boundary and sparse regions, whereas density-aware FPS retains global coverage while selecting more representatives from densely populated normal regions. Consequently, normal test features falling in these high-density regions are more likely to find close memory neighbors, reducing their anomaly scores and suppressing false positives. At the 5\% ratio, DA-Core requires 13.12 ms per electrode sample, compared with 23.39 ms for PatchCore at a 0.20 coreset ratio. As the coreset ratio approaches 1.0, the two strategies become equivalent because all memory candidates are retained. 

\subsubsection{Sensitivity to density weight.}

We also analyze the sensitivity of DA-Core to the density weight $\lambda$ using a fixed coreset ratio of 0.05. Table~\ref{tab:beta_sensitivity} reports the mean and standard deviation of FPR95 over the 10 official splits. Without density weighting, the model obtains 60.4\% FPR95. 
Introducing density awareness consistently reduces FPR95, with the best result achieved at $\lambda=0.7$ (54.3\%). This indicates that an appropriate density weight strikes the best trade-off between maintaining global feature-space coverage and prioritizing representative patterns in dense normal regions.

\begin{table}[t]
\centering
\small
\begin{tabular}{cc}
\toprule
Density weight $\lambda$ & FPR95 $\downarrow$ \\
\midrule
0.0 & 0.604 \textpm\ 0.037 \\
0.3 & 0.571 \textpm\ 0.031 \\
0.4 & 0.582 \textpm\ 0.033 \\
0.5 & 0.568 \textpm\ 0.051 \\
0.6 & 0.562 \textpm\ 0.044 \\
0.7 & \textbf{0.543 \textpm\ 0.057} \\
0.8 & 0.547 \textpm\ 0.053 \\
0.9 & 0.544 \textpm\ 0.048 \\
1.0 & 0.555 \textpm\ 0.051 \\
\bottomrule
\end{tabular}
\caption{Sensitivity analysis of the density weight $\lambda$ using a coreset ratio of 0.05.}
\label{tab:beta_sensitivity}
\end{table}

\section{Conclusion}

We introduced LIBAD, a multimodal industrial anomaly detection benchmark for Li-ion battery electrode manufacturing. The dataset is collected from real roll-to-roll production lines and provides spatially aligned visible-light and X-ray observations, enabling evaluation under an inline-compatible VIS+X-ray inspection setting. LIBAD differs from existing object-centric datasets because electrode patches are highly homogeneous and defect evidence can be modality-dependent, often weak or inconsistent across imaging settings.

Our benchmark results show that existing multimodal anomaly detection methods developed for RGB-3D inspection have limited transferability to LIBAD. Although VIS and X-ray provide complementary evidence, false-positive control remains a key challenge for existing methods, motivating our design. We therefore proposed DA-Core, a density-aware memory-based baseline that improves coreset construction by up-weighting features in dense normal regions during selection. DA-Core achieves better FPR95 than adapted multimodal baselines while preserving the efficiency of memory-based detection. However, despite the relative improvement, the absolute FPR95 remains too high for direct deployment, indicating that LIBAD is still an open and challenging benchmark. 

We expect LIBAD to encourage further study of multimodal IAD in process manufacturing and, together with the proposed method, to contribute to safer battery manufacturing, ultimately benefiting the reliability of electric vehicles and energy storage systems. Future work includes developing fusion strategies that account for modality disagreement, classifying detected anomalies into specific defect categories, and efficient inline deployment, with the longer-term goal of supporting closed-loop process control.

\section*{Acknowledgments}
The authors thank Guangdong Canrd New Energy Technology Co., Ltd. for manufacturing and labeling the electrode samples used in this study. This work received funding from Horizon Europe through the MSCA Doctoral Network RELIANCE, grant no. 101073040, as well as from an internal research grant of Thermo Fisher Scientific (CRC D23016).

\clearpage
\enlargethispage{1\baselineskip}
\bibliography{aaai2027}

\clearpage
\appendix
\section{Appendix}
\setcounter{figure}{0}
\setcounter{table}{0}
\renewcommand{\thefigure}{S\arabic{figure}}
\renewcommand{\thetable}{S\arabic{table}}

\renewcommand{\theHfigure}{S\arabic{figure}}
\renewcommand{\theHtable}{S\arabic{table}}

\subsection*{Note 1: Production Line, Electrode Materials, and Multimodal Acquisition}
\begin{figure}[ht]
 \centering
 \includegraphics[width=\columnwidth]{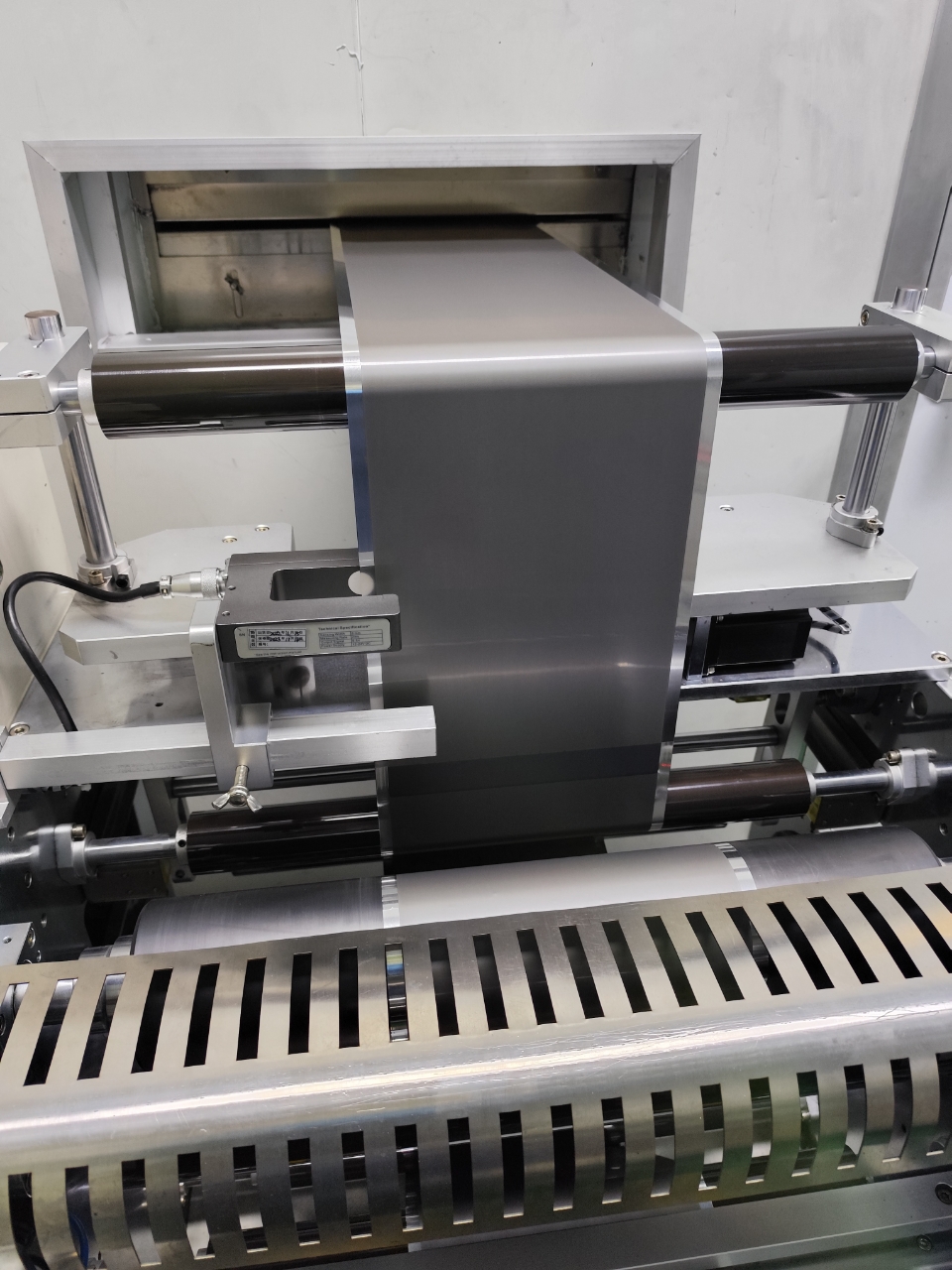}
 \caption{\textbf{Coating module in roll-to-roll electrode manufacturing.} Photograph of the coating section of an industrial roll-to-roll electrode production line, where active material slurry is continuously deposited onto the current collector. This stage represents the upstream process during which localized coating non-uniformities or material-related anomalies may originate and subsequently propagate downstream.}
 \label{fig:coating}
\end{figure}

\begin{figure}[ht]
 \centering
 \includegraphics[width=\columnwidth]{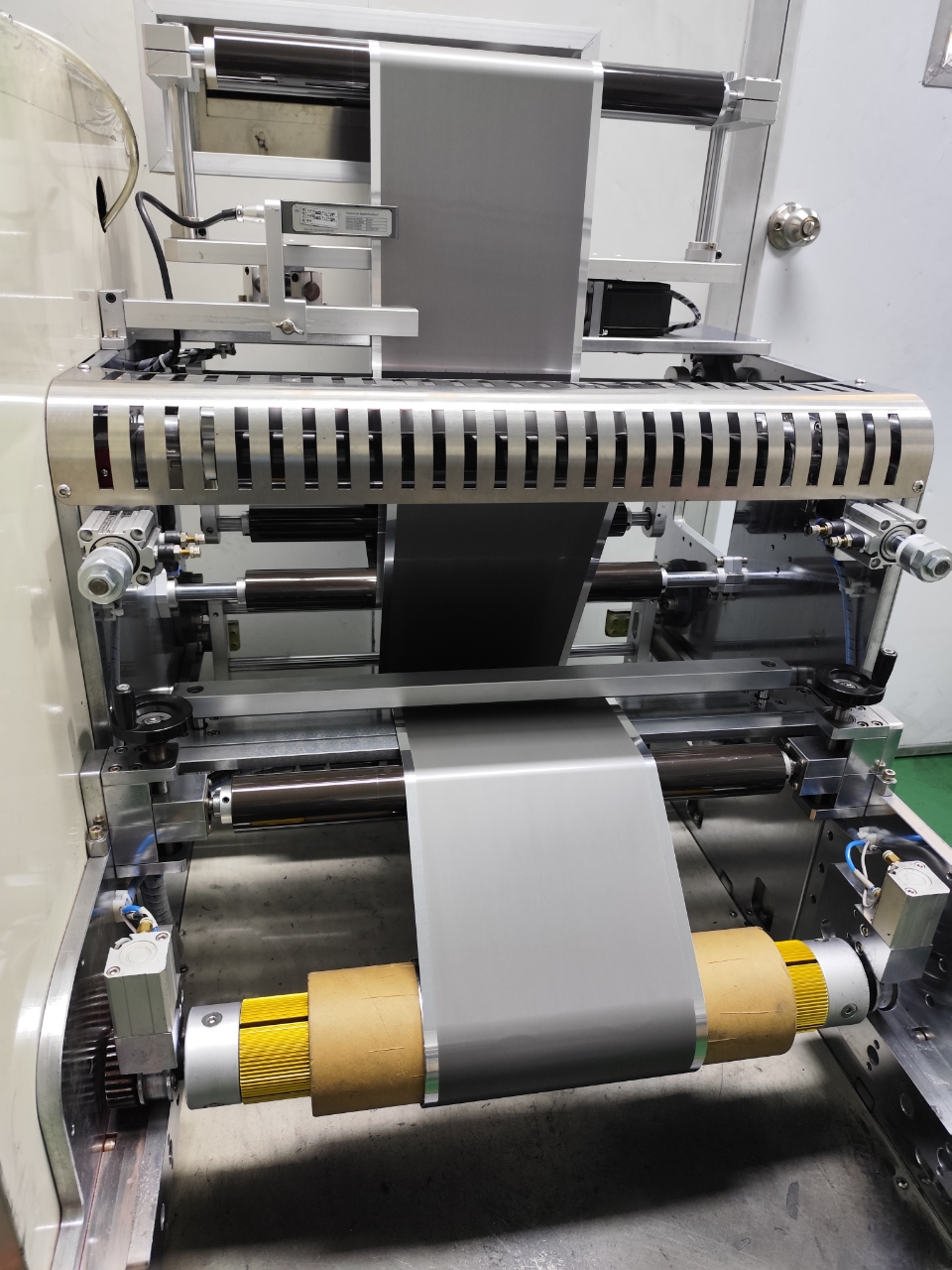}
 \caption{\textbf{Rewinding module in roll-to-roll electrode manufacturing.} Photograph of the rewinding section of the same production line, where the coated electrode web is mechanically guided and rewound.}
 \label{fig:rewinding}
\end{figure}

The dataset used in this study was constructed from electrode samples collected on an automated roll-to-roll lithium-ion battery electrode production line under standard manufacturing operation as shown in Figs.~\ref{fig:coating}, \ref{fig:rewinding}.

LIBAD contains both anode and cathode samples collected across multiple material categories and production conditions. Because the exact active-material formulations of all samples are commercially confidential, their detailed compositions are unavailable. We therefore performed X-ray fluorescence (XRF) analysis to infer the most likely material families from the detected elemental signatures. The resulting labels should be interpreted as approximate material categories rather than confirmed chemical formulations. Samples that could be identified as cathodes but could not be reliably assigned to a specific chemistry were grouped as cathodes with unspecified composition. Among the 744 samples, 313 are anode samples and 431 are cathode samples, accounting for 42.1\% and 57.9\% of the dataset, respectively.

\begin{table}[ht]
\centering
\small
\setlength{\tabcolsep}{7pt}
\begin{tabular}{lrr}
\toprule
Electrode material & \# Samples & Percentage \\
\midrule
Anode & 313 & 42.1\% \\
NMC/NCM cathode & 296 & 39.8\% \\
LFP cathode & 79 & 10.6\% \\
LCO cathode & 4 & 0.5\% \\
Cathode, unspecified & 52 & 7.0\% \\
\midrule
Total & 744 & 100.0\% \\
\bottomrule
\end{tabular}
\caption{Distribution of approximate electrode material categories in LIBAD. Because the exact active-material formulations are commercially confidential, the reported material families are inferred from XRF elemental signatures and should not be interpreted as confirmed chemical compositions. NMC and NCM records are grouped into a single nickel--manganese--cobalt cathode category. Samples that can be identified as cathodes but cannot be reliably assigned to a specific material family are grouped as cathodes with unspecified composition. Percentages are calculated over all 744 electrode samples.}
\label{tab:material_statistics}
\end{table}

\begin{figure}[ht]
 \centering
 \includegraphics[width=\columnwidth]{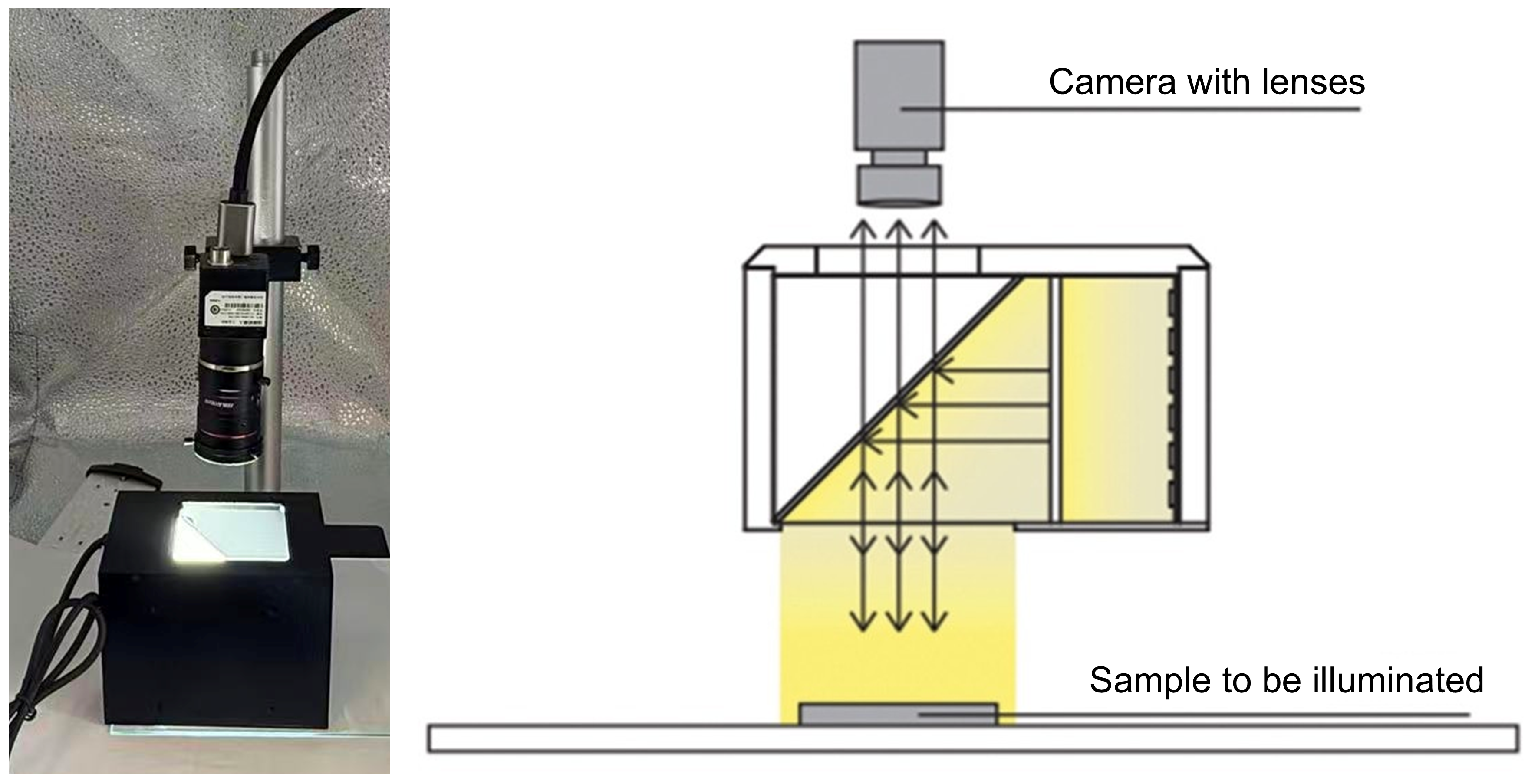}
  \caption{\textbf{Visible-light image acquisition setup and illustration figure of co-axial illumination.} Photograph of the visible-light imaging system used for patch-level data acquisition. The setup consists of an industrial camera combined with a coaxial illumination configuration, enabling consistent surface imaging of both sides of double-coated electrodes under controlled lighting conditions.}
 \label{fig:visaquz}
\end{figure}

\begin{figure}[ht]
 \centering
 \includegraphics[width=\columnwidth]{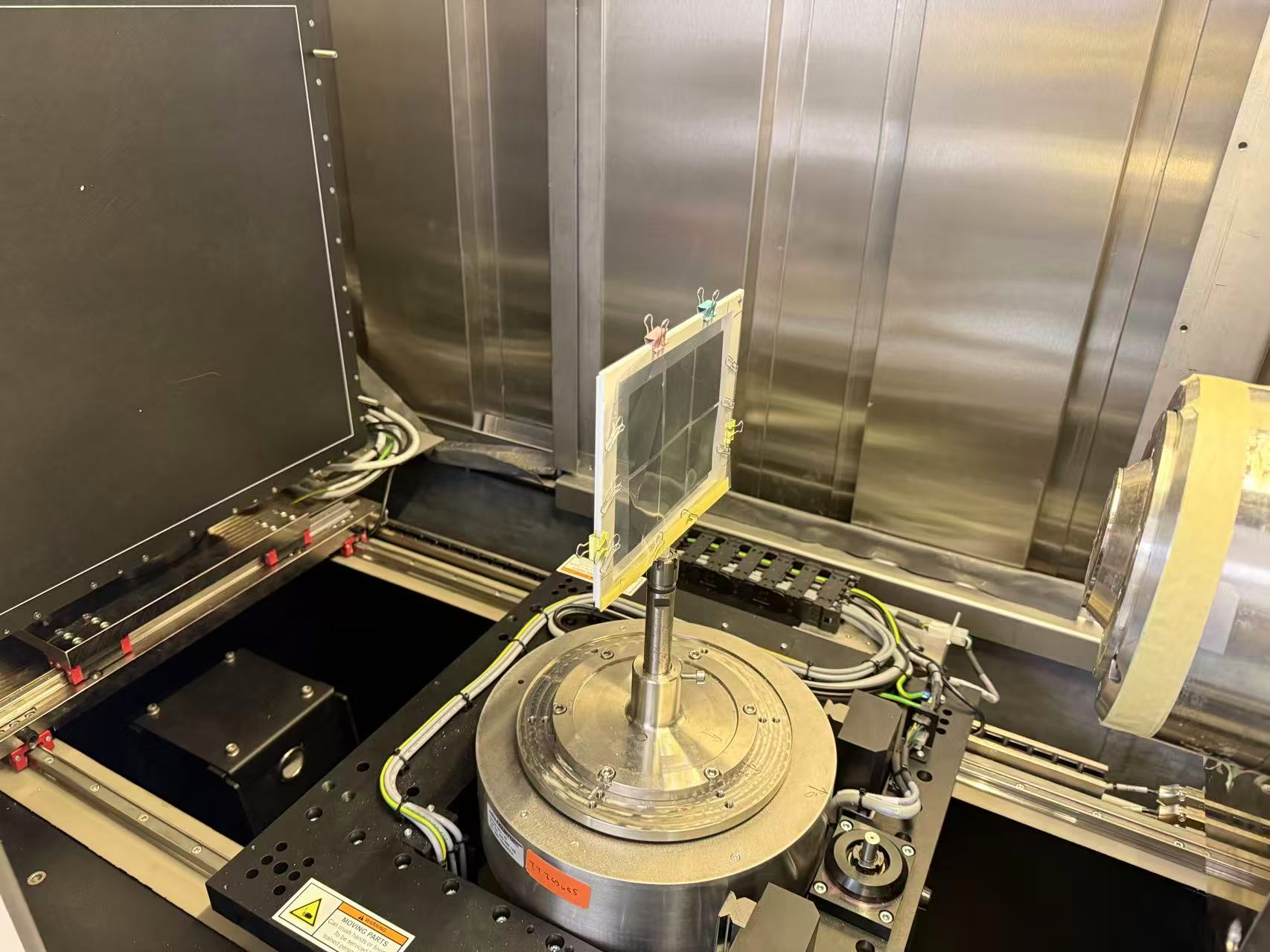}
  \caption{\textbf{High-resolution X-ray radiography acquisition setup.} Photograph of the high-resolution X-ray imaging system used for offline characterization of excised electrode patches. This setup provides subsurface contrast at a spatial resolution exceeding that of inline-compatible X-ray inspection and was used exclusively for post-extraction analysis.}
 \label{fig:XrayHaquz}
\end{figure}

Following physical extraction from the electrode web, multimodal images were acquired for each electrode patch using separate visible-light and X-ray imaging systems. Visible-light images were obtained under controlled illumination conditions to capture surface features from both sides of the double-coated electrodes (Fig.~\ref{fig:visaquz}). High-resolution X-ray radiographs were acquired offline to provide subsurface contrast that is not accessible through optical inspection (Fig.~\ref{fig:XrayHaquz}). In addition, inline-compatible low-resolution X-ray measurements were acquired under throughput-constrained conditions representative of roll-to-roll manufacturing. Further details of the acquisition system are omitted due to commercial confidentiality.

\subsection*{Note 2: Sample-region Extraction and Multimodal Registration}

\begin{figure}[ht]
    \centering
    \includegraphics[width=\columnwidth]{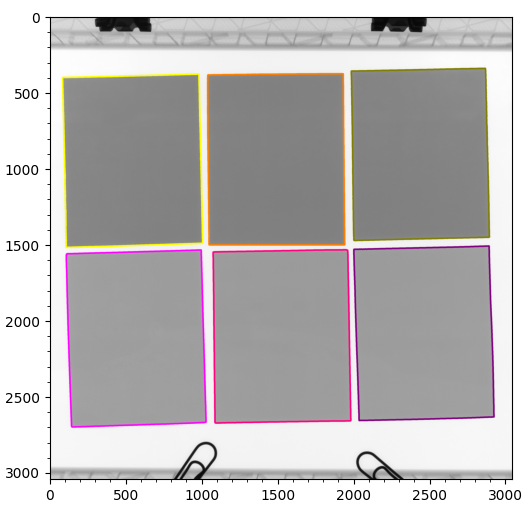}
    \caption{\textbf{Sample-region detection and orientation estimation.}
    Example raw X-rayH acquisition image containing six electrode patches. The overlaid contours indicate the automatically detected sample boundaries. These regions are used to estimate the in-plane orientation of each patch, correct its rotation, and crop it as an individual sample before multimodal registration. The same sample-region extraction principle is applied to all imaging modalities, with modality-specific preprocessing where required.}
    \label{fig:sample_extraction}
\end{figure}

Raw acquisition images may contain multiple electrode patches within a single field of view. Before multimodal registration, we therefore first detect and extract each individual sample region. For each modality, the outer contour of every electrode patch is detected after modality-specific intensity preprocessing. The in-plane orientation of each valid sample is estimated from its detected region. Each patch is then rotated to a common orientation and cropped as an independent sample image. Figure~\ref{fig:sample_extraction} shows an example X-rayH acquisition image containing six electrode samples and the detected contours used for orientation correction and sample extraction.

\begin{figure}[ht]
    \centering
    \includegraphics[width=\columnwidth]{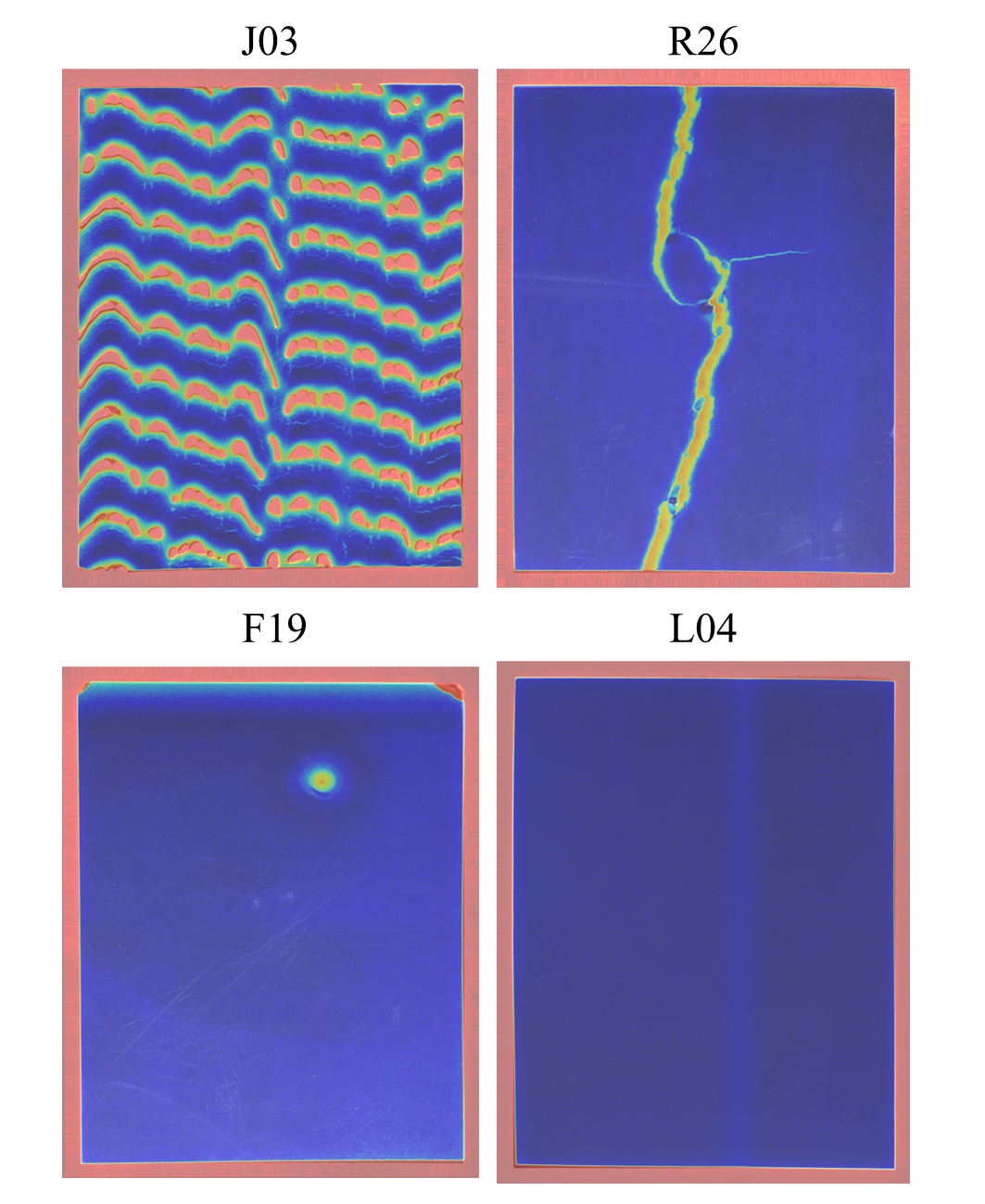}
    \caption{\textbf{Qualitative visualization of VIS--X-rayH registration.}
    Representative registration results for four electrode samples. In each panel, the VIS image is shown in grayscale, and the registered X-rayH radiograph is overlaid as a semitransparent heatmap. The agreement between electrode boundaries and other corresponding spatial structures demonstrates the effectiveness of the multimodal registration procedure.}
    \label{fig:registration_overlay}
\end{figure}

After sample extraction and orientation correction, images from different modalities are transformed into a common spatial coordinate system using the registration procedure described in the main paper. VIS is used as the fixed reference, and the corresponding X-ray image is transformed into the VIS coordinate system. Figure~\ref{fig:registration_overlay} presents representative registration overlays for four samples. In each panel, the VIS image is shown in grayscale, while the registered X-rayH radiograph is displayed as a semitransparent heatmap. The close agreement between electrode boundaries and other spatially corresponding structures provides qualitative evidence of reliable cross-modal alignment. These examples complement the landmark-based registration error reported in the main paper.

\FloatBarrier
\subsection*{Note 3: Dataset Re-annotation and Representative Defect Examples}

\begin{figure}[ht]
 \centering
 \includegraphics[width=0.9\columnwidth]{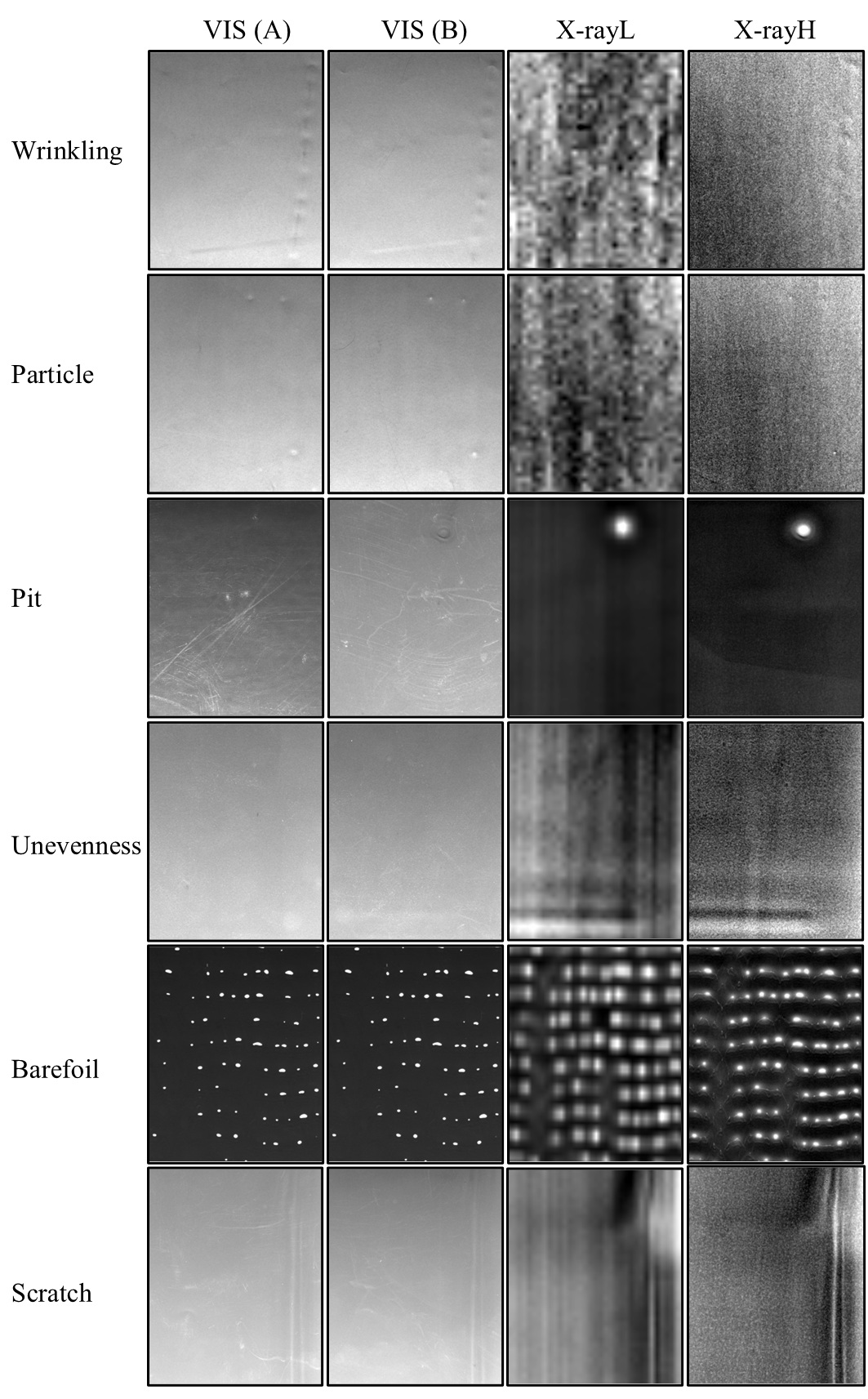}
 \caption{Examples of Wrinkling, Particle, Pit, Unevenness, Barefoil, and Scratch defects (rows) shown across visible-light imaging of side A (VIS A), visible-light imaging of side B (VIS B), low-resolution inline X-ray (X-rayL), and high-resolution X-ray radiography (X-rayH) (columns).}
 \label{fig:defects1}
\end{figure}

\begin{figure}[ht]
 \centering
 \includegraphics[width=0.9\columnwidth]{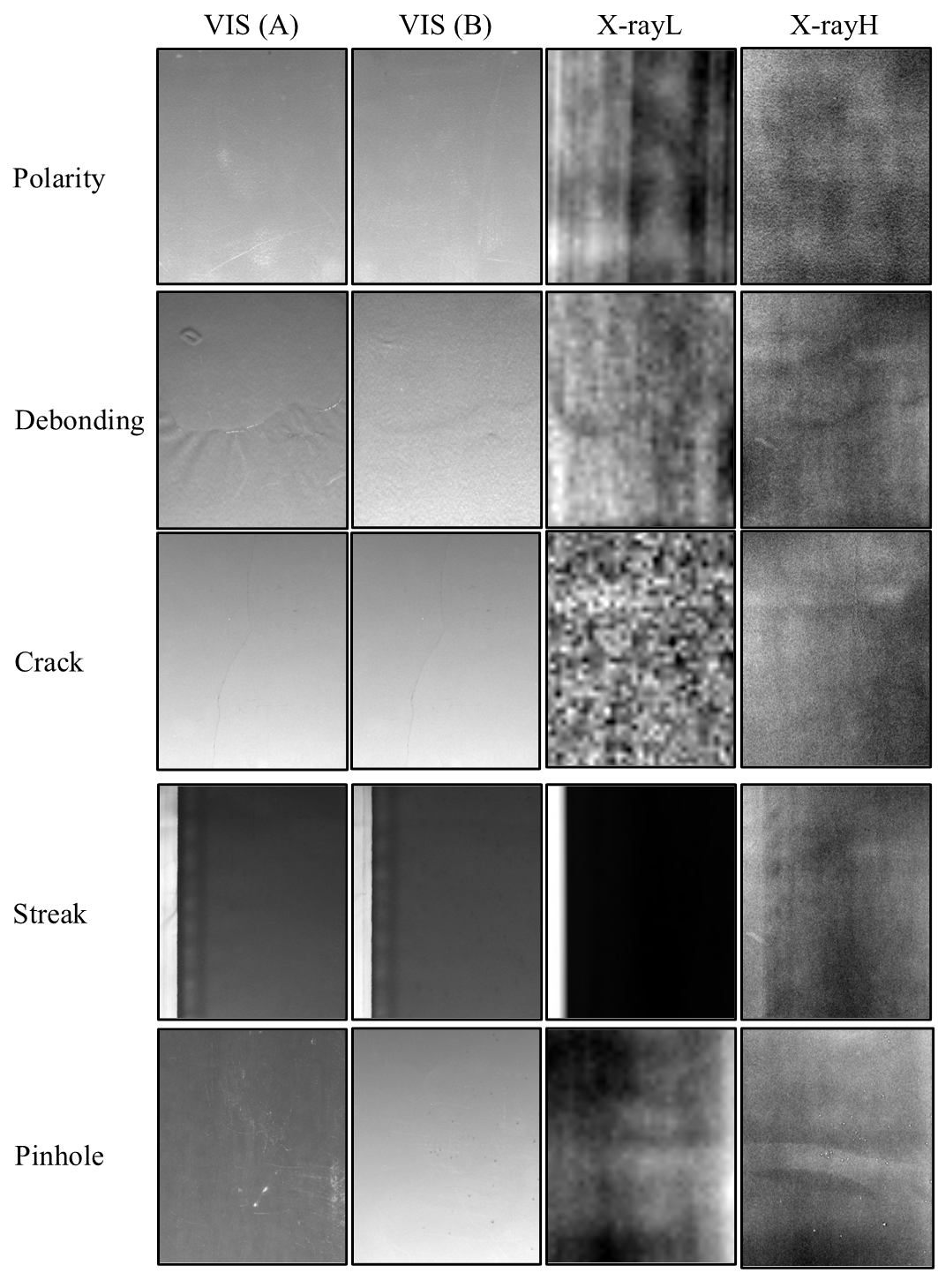}
 \caption{Examples of Polarity, Debonding, Crack, Streak and Pinhole defects (rows) shown across visible-light imaging of side A (VIS A), visible-light imaging of side B (VIS B), low-resolution inline X-ray (X-rayL), and high-resolution X-ray radiography (X-rayH) (columns).}
 \label{fig:defects2}
\end{figure}

After completing high-resolution X-ray acquisition and multimodal registration, we performed an exploratory analysis of the complete image collection. During this analysis, we found that some samples had been assigned labels under the original taxonomy, whereas their X-ray images revealed subsurface density irregularities inconsistent with those labels. We therefore conducted a re-annotation of the entire dataset using the registered visible-light and X-ray observations. Because manufacturing-data access and sampling practices are governed by customer confidentiality and operational constraints, certain quantities, such as the absolute prevalence of individual defect types and the exact sampling fraction relative to total production, cannot be disclosed or retrospectively quantified. LIBAD is therefore intended to support comparative analysis of defect observability across sensing modalities under realistic industrial conditions, rather than to represent production-level defect prevalence. Representative examples of 11 defect classes are shown in Figs.~\ref{fig:defects1}--\ref{fig:defects2}. 

\FloatBarrier
\subsection*{Note 4: Modality-dependent Defect Visibility}

Defect visibility was assessed through manual inspection of the modality images. Before inspection, each image was percentile-rescaled to enhance defect contrast and suppress the influence of extreme-intensity noise. A defect was considered visible in a modality when its anomalous region could be visually localized from that modality after rescaling. For samples in which the spatial correspondence between modalities was difficult to assess directly, the mutual-information score obtained during multimodal registration was additionally consulted as a reference for evaluating the reliability of the cross-modal correspondence.

LIBAD exhibits clear modality-dependent defect visibility. Among the 383 anomalous samples, 312 could be localized from VIS and 341 could be localized from X-rayH. Of these, 270 samples were visible in both modalities, 42 were visible only in VIS, and 71 were visible only in X-rayH. As summarized in Table~\ref{tab:defect_visibility}, 29.5\% of all anomalous samples were visible in only one modality, including 11.0\% that were visible only in VIS and 18.5\% that were visible only in X-rayH. This modality-selective subset provides direct evidence of cross-modal anomaly inconsistency in LIBAD and demonstrates that neither modality alone provides complete defect coverage.

\begin{table}[h]
\centering
\small
\setlength{\tabcolsep}{8pt}
\begin{tabular}{lrr}
\toprule
Defect visibility & \# Samples & Percentage \\
\midrule
VIS only & 42 & 11.0\% \\
X-rayH only & 71 & 18.5\% \\
Visible in both modalities & 270 & 70.5\% \\
\midrule
Total & 383 & 100.0\% \\
\bottomrule
\end{tabular}
\caption{Modality-dependent visibility of anomalous samples in LIBAD, determined by manual inspection after percentile-based intensity rescaling. A total of 113 anomalous samples, corresponding to 29.5\% of all defects, are visible in only one of the two modalities.}
\label{tab:defect_visibility}
\end{table}

\subsection*{Note 5: Effect of Backbone Selection}

\begin{table*}[ht]
\centering
\small
\setlength{\tabcolsep}{5.5pt}
\begin{tabular}{lccc ccc ccc}
\toprule
& \multicolumn{3}{c}{DINOv3}
& \multicolumn{3}{c}{DINOv2}
& \multicolumn{3}{c}{DINOv1} \\
\cmidrule(lr){2-4}
\cmidrule(lr){5-7}
\cmidrule(lr){8-10}
Method
& AUROC $\uparrow$
& AUPR $\uparrow$
& FPR95 $\downarrow$
& AUROC $\uparrow$
& AUPR $\uparrow$
& FPR95 $\downarrow$
& AUROC $\uparrow$
& AUPR $\uparrow$
& FPR95 $\downarrow$ \\
\midrule
CFM
& \textbf{0.834}
& \textbf{0.947}
& \textbf{0.618}
& 0.765
& 0.925
& 0.802
& 0.737
& 0.912
& 0.817 \\

M3DM
& \textbf{0.862}
& \textbf{0.957}
& \textbf{0.616}
& 0.852
& 0.956
& 0.661
& 0.786
& 0.932
& 0.778 \\

CFR
& \textbf{0.833}
& \textbf{0.948}
& \textbf{0.681}
& 0.786
& 0.932
& 0.789
& 0.754
& 0.917
& 0.775 \\

G$^{2}$SF
& \textbf{0.852}
& \textbf{0.953}
& \textbf{0.595}
& 0.843
& 0.952
& 0.667
& 0.755
& 0.920
& 0.750 \\
\bottomrule
\end{tabular}
\caption{Effect of the backbone version on adapted multimodal anomaly detection methods under the VIS+X-rayL setting. All methods use the same evaluation protocol, and results are averaged over the 10 official splits. The best result for each method and metric is highlighted in bold. Replacing the original DINOv1 or DINOv2 backbone with DINOv3 consistently improves anomaly-ranking performance and reduces FPR95.}
\label{tab:backbone_comparison}
\end{table*}

Table~\ref{tab:backbone_comparison} compares DINOv1, DINOv2, and DINOv3 as the feature backbone for the adapted multimodal baselines. DINOv3 consistently achieves the best results across all four methods and all three evaluation metrics. The improvement is particularly evident in FPR95. For example, replacing DINOv1 with DINOv3 reduces FPR95 from 81.7\% to 61.8\% for CFM, from 77.8\% to 61.6\% for M3DM, from 77.5\% to 68.1\% for CFR, and from 75.0\% to 59.5\% for G$^{2}$SF. We therefore use DINOv3 for all adapted baselines in the main benchmark to provide a stronger and more consistent comparison.

\subsection*{Note 6: Experimental Environment and Implementation Details}

All experiments are conducted on a workstation running Ubuntu 22.04, equipped with an NVIDIA RTX PRO 6000 Blackwell GPU with 96~GB of GPU memory and a 300~W power limit, an Intel Xeon w5-2455X CPU, and 256~GB of system memory. The software environment uses Python 3.12.13, PyTorch 2.12.1, and CUDA 13.0. Computational resource allocation and job scheduling are managed using Slurm 25.05.0. The batch size is set to 16 for both training and inference-time evaluation.

The frozen DINOv3 backbone is evaluated using BF16 precision. Memory-bank distance matrices and nearest-neighbor distances are computed in FP32, with TF32 disabled to ensure consistent numerical precision across all methods. The OCSVM-based decision-fusion module is implemented using scikit-learn's SGDOneClassSVM and executed on the CPU. All methods are evaluated under the same hardware, software, numerical-precision, and input-resolution settings.

For runtime evaluation, the first inference batch is treated as a warm-up and excluded from the runtime statistics. CUDA synchronization is applied immediately before and after each timed batch, and wall-clock time is measured using a high-resolution timer. The reported time per electrode sample is calculated by dividing the accumulated runtime of all measured batches by the corresponding number of samples. Each runtime experiment is repeated three times, and the mean runtime is reported.

The timed section includes input preprocessing after data loading, input transfer and preparation, backbone feature extraction, modality-specific feature processing, memory-bank nearest-neighbor matching, image-level anomaly scoring, OCSVM-based decision fusion, and the post-processing required to produce the final anomaly scores. Data loading, evaluation-metric computation, visualization, and file I/O are excluded.

\end{document}